\pdfoutput=1

\documentclass[11pt]{article}

\usepackage[]{emnlp22}

\usepackage{times}
\usepackage{latexsym}
\usepackage{graphicx}
\usepackage[T1]{fontenc}
\usepackage{cleveref}

\usepackage[utf8]{inputenc}

\usepackage{microtype}

\usepackage{csvsimple} 
\usepackage{subfig}
\usepackage{caption}
\usepackage{tablefootnote}
\usepackage{float}
\usepackage{pdfpages}

\usepackage{tabularx}
\usepackage{booktabs, ragged2e}

\newcommand{\dalle}{\texttt{DALLE-2}}
\newcommand{\dallemini}{\texttt{DALLE-mini}}
\newcommand{\stable}{\texttt{Stable-diffusion}}

\usepackage[export]{adjustbox}

\title{DALLE-2 is Seeing Double: Flaws in Word-to-Concept Mapping in Text2Image Models
}

 \author{Royi Rassin$\footnotemark[1]$ \textsuperscript{{ \normalfont 1}}, Shauli Ravfogel$\thanks{~~Equal contribution.}$ \textsuperscript{{ \normalfont 1,2}} \, Yoav Goldberg\textsuperscript{\normalfont 1,2} \\
\textsuperscript{1}Bar-Ilan University \, \textsuperscript{2}Allen Institute for Artificial Intelligence \\
  {\tt\{\href{mailto:rassinroyi@gmail.com}{rassinroyi},
  \href{mailto:shauli.ravfogel@gmail.com}{shauli.ravfogel}, \href{mailto:yoav.goldberg@gmail.com}{yoav.goldberg}\}@gmail.com} 
  }

\begin{document}
\maketitle
\begin{abstract}
We study the way \dalle\ maps symbols (words) in the prompt to their references (entities or properties of entities in the generated image). We show that in stark contrast to the way human process language,  \dalle\ does not follow the constraint that each word has a single role in the interpretation, and sometimes re-uses the same symbol for different purposes. We collect a set of stimuli that reflect the phenomenon: we show that \dalle\ depicts both senses of nouns with multiple senses at once; and that a given word can modify the properties of two distinct entities in the image, or can be depicted as one object and also modify the properties of another object, creating a semantic leakage of properties between entities. Taken together, our study highlights the differences between \dalle\ and human language processing and opens an avenue for future study on the inductive biases of text-to-image models.
\end{abstract}
 
\section{Introduction}

Large diffusion-based text-to-image models, such as \dalle\ \citep{dalle2}, generate visually compelling images that condition on input texts. Yet, the extent to which such models capture properties of human language, such as its compositional structure, has been doubted \citep{prep_analysis}.

\begin{figure}[]
    \centering
    \includegraphics[width=0.75\columnwidth]{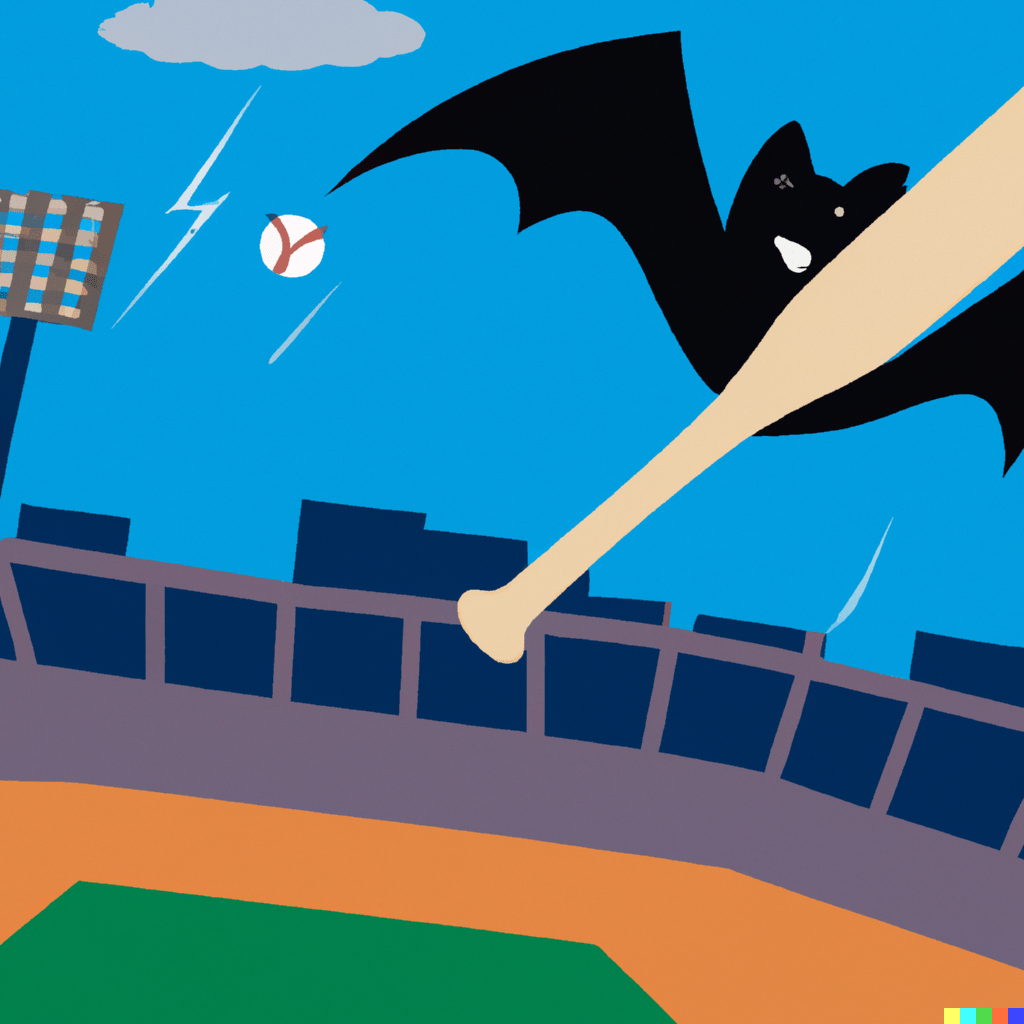}
    \caption{The word \emph{bat} is realized as two entities given the prompt \emph{a bat is flying over a baseball stadium}.}
\label{fig:flying-bat}
\end{figure}

A very basic property in the interpretation of natural language utterances is that each word has a specific role in the interpretation, and there is a one-to-one mapping between symbols and roles. While symbols---as well as sentence structures---may be ambiguous, after an interpretation is constructed this ambiguity is already resolved. For example, while the symbol \emph{bat} in \emph{a flying bat} can be interpreted as either a wooden stick or an animal, our possible interpretations of the sentence are either of a \emph{flying wooden stick} or \emph{a flying animal}, but never both at the same time.
Once the word \emph{bat} has been used in the interpretation to denote an object (for example a wooden stick), it cannot be re-used to denote another object (an animal) in the same interpretation. Similarly, in \emph{a fish and a gold ingot}, the word \emph{gold} is used as a modifier of \emph{ingot}.\footnote{We use the word "modifier" to refer to any word that influences the way another word is depicted in the output.} Once it is used, it cannot be re-used to modify another object in the same interpretation, and also cannot be used as a standalone object.\footnote{Note that in some cases a single word can be used to modify several objects, for example, \emph{a gold fish and ingot} can be interpreted as \emph{a gold fish and a gold ingot}. These cases manifest in very specific syntactic configurations, and are well documented in linguistics. Indeed, the linguistic analysis of such cases are either based on mechanisms such as ``duplication'' of the word, or allude to a deeper version of the sentence in which the word appeared twice, and was dropped before production. While the reality behind these linguistic theories cannot be proven, the appeal to use such mechanisms as ``duplication'' or ``deletion" (before realization) as explanations to the phenomena, highlights the naturalness of the ``single use per symbol'' principle.} Some linguists refer to this property as \emph{resource sensitivity} \cite{salvucci2009toward}.

\begin{figure}[]
    \centering
    \includegraphics[width=0.75\columnwidth]{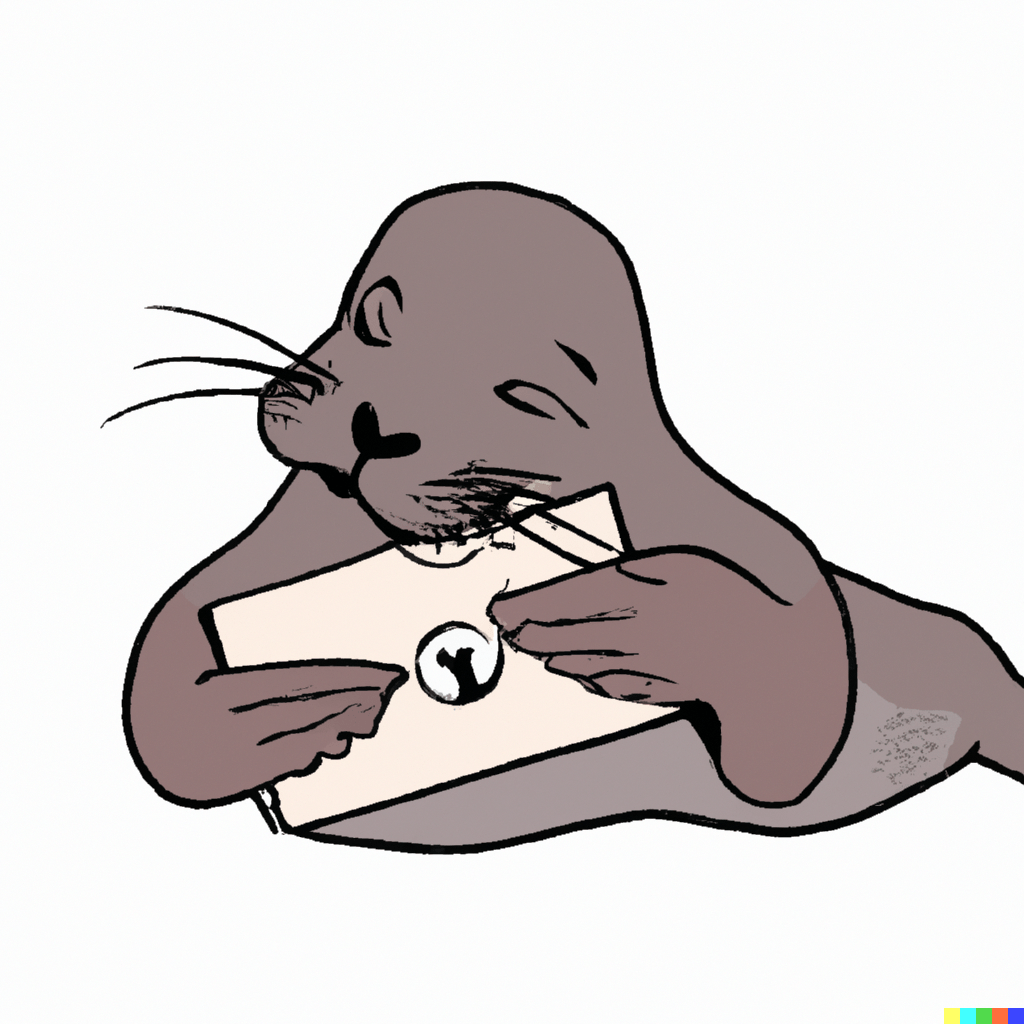}
    \caption{The word \emph{seal} is realized both as an object and as a modifier of another word (\emph{letter}) in the prompt \emph{a seal is opening a letter}, resulting in \emph{concept leakage}. The prompt tends to generate \emph{sealed} letters, while minimally-different prompts that do not contain the noun \emph{seal} do not.}
    \label{fig:seal-letter}
\end{figure}

\begin{figure}[h!]         
    \begin{centering}
    \includegraphics[width=0.20\textwidth]{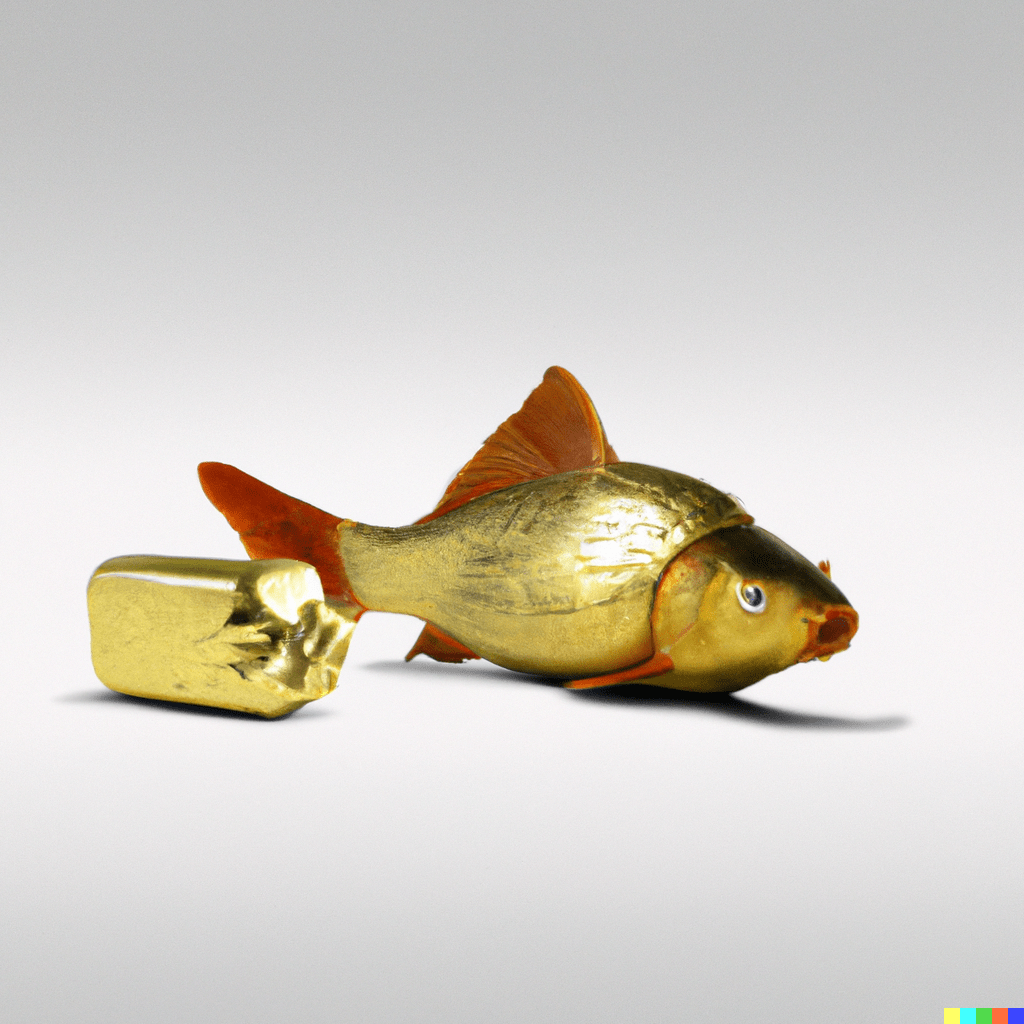}
    \includegraphics[width=0.20\textwidth]{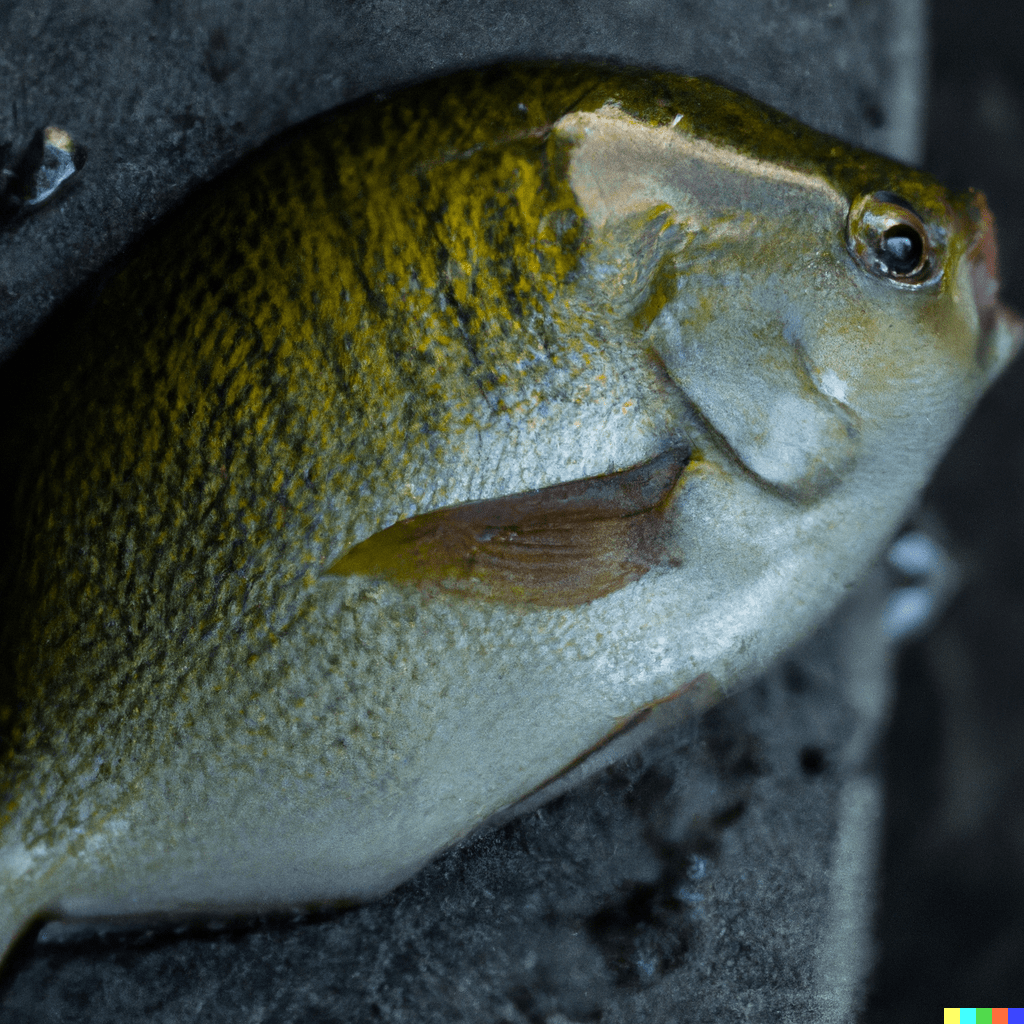}
    \end{centering}
    \\\noindent\textbf{Main} (left): \emph{a \textbf{fish} and a \textbf{\underline{gold}} \textbf{ingot}}. \\\noindent\textbf{Control} (right): \emph{a fish and an ingot}.\\[0.5em]
    \label{fig:fish_example}
\end{figure}

We show evidence that---in stark contrast to humans---the text-to-image model \dalle\ \citep{dalle2} does not respect this constraint. Indeed, we show that \dalle\ exhibits the following behaviors that are inconsistent with the single-role principle:
\begin{itemize}
    \item \textbf{A word or phrase is interpreted as two distinct (concurrently-incompatible) entities} and is consequently realized as multiple objects in the same scene. For instance, the prompt \emph{a bat is flying over a stadium} is mapped to an image showing a baseball bat alongside the flying mammal (\cref{fig:flying-bat}).
    \item \textbf{A word is interpreted as a modifier of two different entities},\footnote{While not under a syntactic configuration that allows such duplication.} and is consequently realized as a property of multiple objects in the scene. For instance, the prompt \emph{a fish and a gold ingot} is mapped to an image showing a gold ingot as well as a goldfish (\cref{fig:fish_example}).
    \item \textbf{A word is interpreted simultaneously as an entity and as a modifier of a different entity} and is consequently realized as an object in the scene as well as as a property of another object in the same scene. For instance, the prompt \emph{a seal is opening a letter} is mapped to an image showing a seal holding a \emph{sealed} letter (\cref{fig:seal-letter}). 
\end{itemize}

Visually, the cases we highlight map to one of two failure modes from the perspective of the image designer / user: (a) sense-ambiguous words are hard to isolate, and the resulting images often exhibit the unintended sense together with the intended one (\emph{homonym duplication}). (b) visual properties of one object in the image "leak" into other objects in the image (\emph{concept leakage}).\footnote{Importantly, we note that both phenomena rarely occur when explicit specification is provided, e.g, the prompt \textit{a fruit bat is flying over a stadium} generates only the flying mammal. See \citet{hutchinson2022underspecification} for a discussion on under-specification in text-to-image generation.}

We also observe some \emph{second order effects}, where a hypernym of the modifier word, or an implicit description of it, also triggers the duplication effect.

Taken together, the phenomena we examine provides evidence for limitations in the linguistic ability of \dalle\, and opens avenues for future research that would uncover whether those stem from issues with the text encoding, the generative model, or both. More generally, the proposed approach can be extended to other scenarios where the decoding process is used to uncover the inductive bias and the shortcomings of text-to-image models. 

\section{Previous Work}
Promptly after the introduction of \dalle\ \citep{dalle2}, analyses of the system began to surface \citep{gary_analysis, prep_analysis}.
\citet{prep_analysis} showed that \dalle\ does not understand spatial relations.
\citet{gary_analysis} 
reported that if an object in the prompt is said to have some property, then the image will likely show that property somewhere, however, not necessarily on the correct object. We take this analysis a step further, showing that the same property sometimes simultaneously modifies several unrelated objects. \citet{dalle2} allude to the phenomena we analyze: they mention the issues with binding separate attributes to separate objects, and hypothesize that it is because CLIP \citep{clip} embeddings do not explicitly bind attributes to objects. Accordingly, they observe that that reconstructions from the decoder often mix up attributes and objects. Concurrent to this work, \citet{hutchinson2022underspecification} discuss under-specification in text-to-image generation, a condition which we find to be associated with some of the behaviors we identify; and \citet{anonymous2023trainingfree} discuss several cases of ``leakage", and propose an intervention in the decoding process that is aimed to mitigate them.


\section{Methdos}
\label{Methodology}

\paragraph{Stimuli} We construct linguistic stimuli (prompts) which evoke a behavior that is inconsistent with the single-use-per-symbol axiom.
For the first case (words interpreted as two entities), we use \emph{homonyms}, words that have two distinct senses. \dalle\ often generates two objects, one corresponding to each of the senses. For the modification cases, we construct the following prompts. For the case of a word that is interpreted as a modifier of two entities, we include prompts with two entities $e_1$ and $e_2$ and a modifier word $m$ that is semantically compatible with both of them, but is only syntactically modifying the second one (For instance, ``A \emph{fish}$_{e_1}$ and a \emph{gold}$_{m}$ \emph{ingot}$_{e_2}$"). For the case of a word that is simultaneously interpreted as an entity as well as a modifier, we construct stimuli that contain two entities $e_1$ and $e_2$, such as that one of them is semantically associated with the other noun---for instance, ``classic \emph{butter}$_{e_1}$ and \emph{peanuts}$_{e_2/m}$". 

The complete list of stimuli is presented in tables \ref{app:homonyms}, \ref{app:two_properties} and \ref{app:object-to-property} in \cref{app:data}, together with samples of the generated images.

\paragraph{Control Stimuli}
In order to argue for the existence of a leakage of property $P$ between two entities $e_1$ and $e_2$, we need to assess whether \dalle\ does not tend to depict $e_2$ with that property, regardless of $e_1$. For instance, if the default \emph{groceries} for \dalle\, in different contexts is a groceries basket, it is hard to argue that it is the presence of the word \emph{basket} in ``a \emph{basketball} near \emph{groceries}" that elicited \dalle\ to generate a basket. With this in mind, we generate a set of control prompts, which are minimally-different variations over the challenge prompts, that were built so as to prevent the possibility of a leakage by changing either $n_1$ or $n_2$ alone, depending on which change can prevent the leakage in a given prompt. The full set of controls are also detailed in tables \ref{app:two_properties} and \ref{app:object-to-property}.

\section{Results}

We generate 12 images per stimulus, and report the average over all stimuli and images.

\subsection{A single word realized as multiple entities}
\begin{figure}[h!]
\begin{centering}
\includegraphics[width=0.22\textwidth]{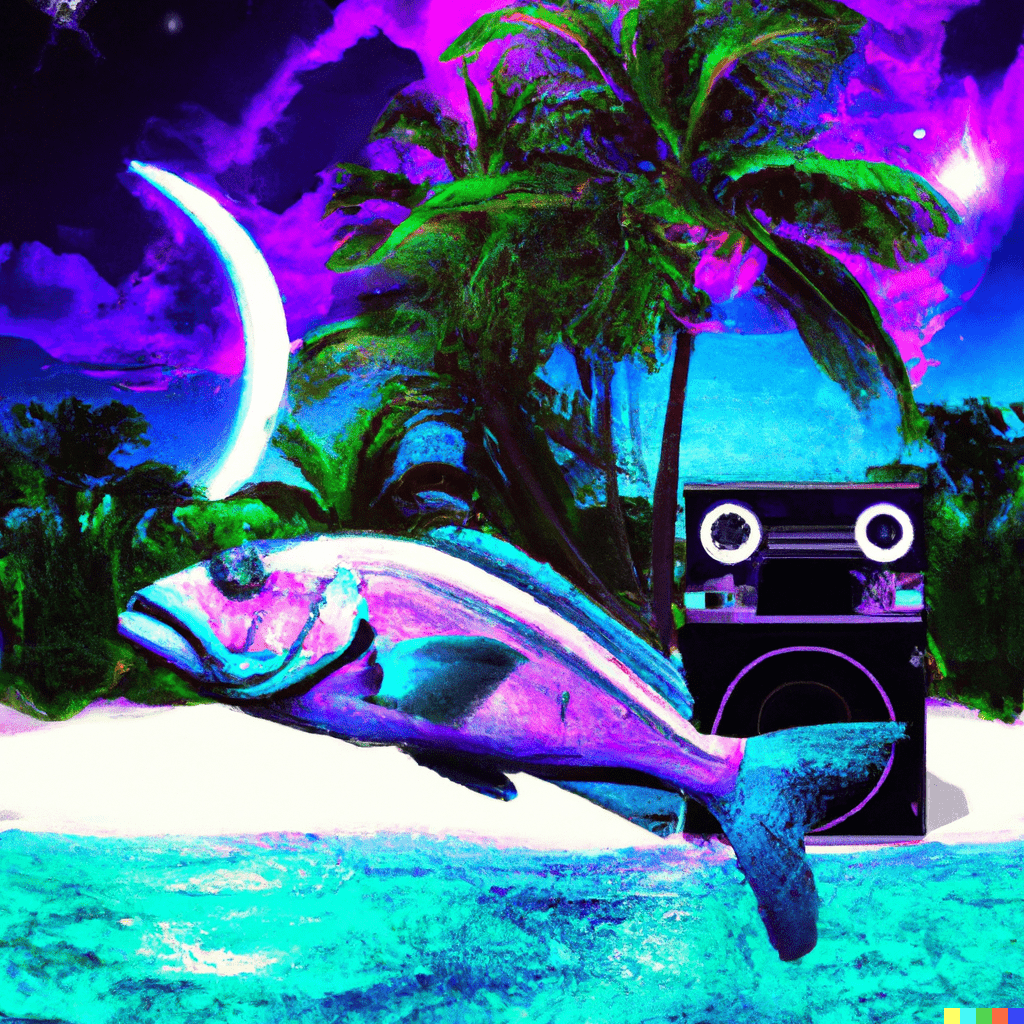}
\includegraphics[width=0.22\textwidth]{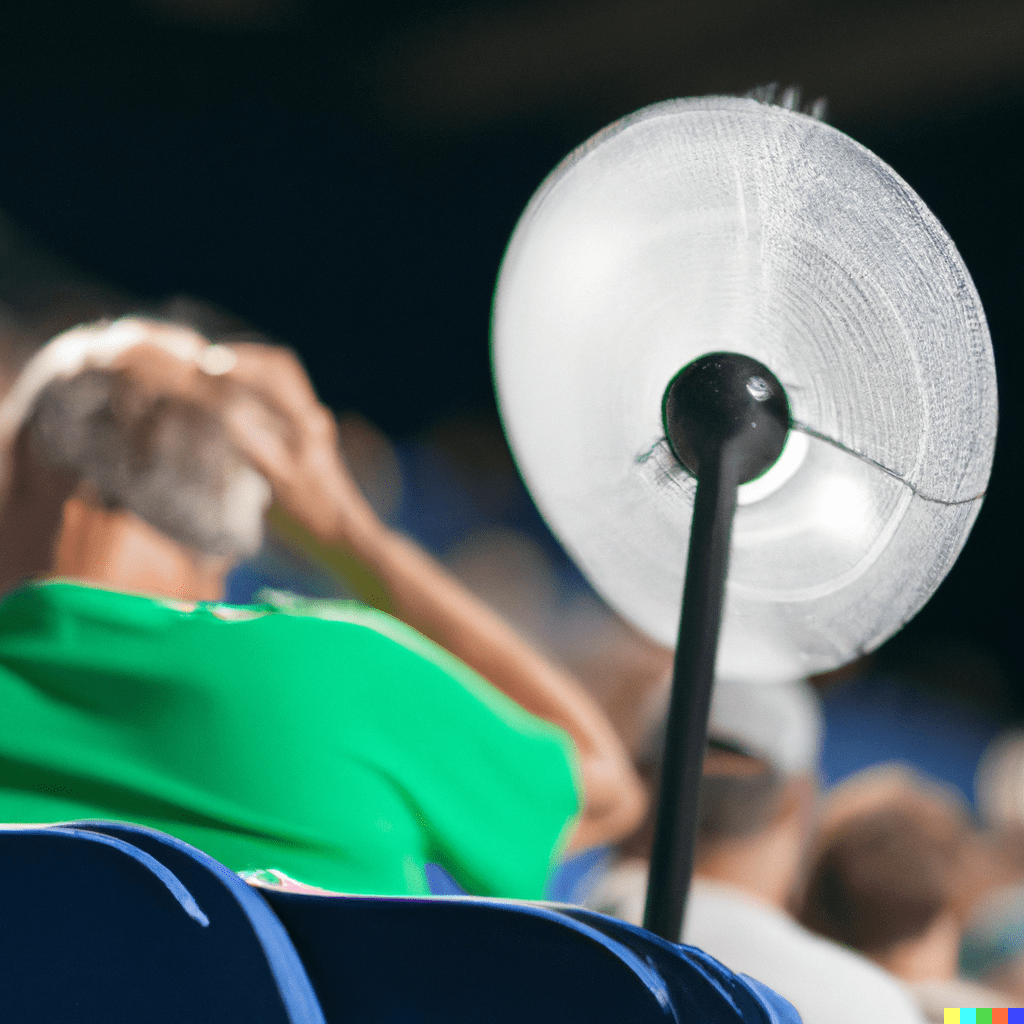}\\
\end{centering}
\noindent left: \emph{A bass lounging in a tropical resort in space, vaporwave}. \\\noindent right: \emph{a fan at a hot sport event}.\\[0.5em]
\label{fig:fan_example}
\end{figure}
\vspace{-0.1cm}


We have 17 stimuli that elicit \dalle\ to assign a single word two roles. Homonym duplication occurred in 80.3\% of the 216 images. We found that, out of context, the vast majority of words are biased towards a particular sense, specifically, out of our 17 homonyms, only \textit{bass} and \textit{date} reveal homonym duplication without added context. Using \textit{bass} as a prompt will\footnote{Using just a homonym as a prompt does not consistently elicit the phenomenon as a prompt that includes context with the homonym.} return an image of a \textit{seabass} and a \textit{guitar bass} or a \textit{speaker bass}, and for \textit{date}, \dalle\ will return a romantic couple, holding together a date (the fruit). Most of the homonym duplication prompts were created by including context related to the non-default role of the homonym. For instance, for the homonym \textit{fan}, \dalle\ is biased towards the cooling device, but if we are to add context that is related to the other possible role a \textit{fan} can have (a sports fan), and that still applies towards the cooling device, we will get both: \textit{A fan at a hot sports event} (top of \cref{fig:fan_example}).

\subsection{A single word realized as a modifier of multiple entities}
\vspace{-0.5cm}

\vspace{-0.5cm}

Given a pair of entities \textit{($e_1$, $e_2$}), we observe that the underspecified entity, $e_2$ is depicted by \dalle\ with properties from $e_1$: in the prompt \textit{A fish and a gold ingot}, \textit{gold} modifies \textit{fish}, and the image consistently contains a golden fish.\footnote{Interestingly, The golden fish also resemble goldfish, suggesting that priming effect also extends to compounds, even though the word \textit{gold} in \textit{gold ingot} is a part of a phrase with a compositional structure, while ``goldfish" is not compositional.} This is likely because \textit{fish} is an underspecified noun and \textit{gold} is an amicable property to fish. In the control, \textit{a fish and an ingot}, golden fish do not appear at all in the output.

For the two-properties case, we have 6 stimuli-control pairs. The output of the stimuli prompts display the shared property in 97.2\% of the cases, while the control prompts show that property in only 15.2\% of the cases.

\subsection{A single word is realized as an entity and as a modifier of another entity}
\begin{centering}
\includegraphics[width=0.22\textwidth]{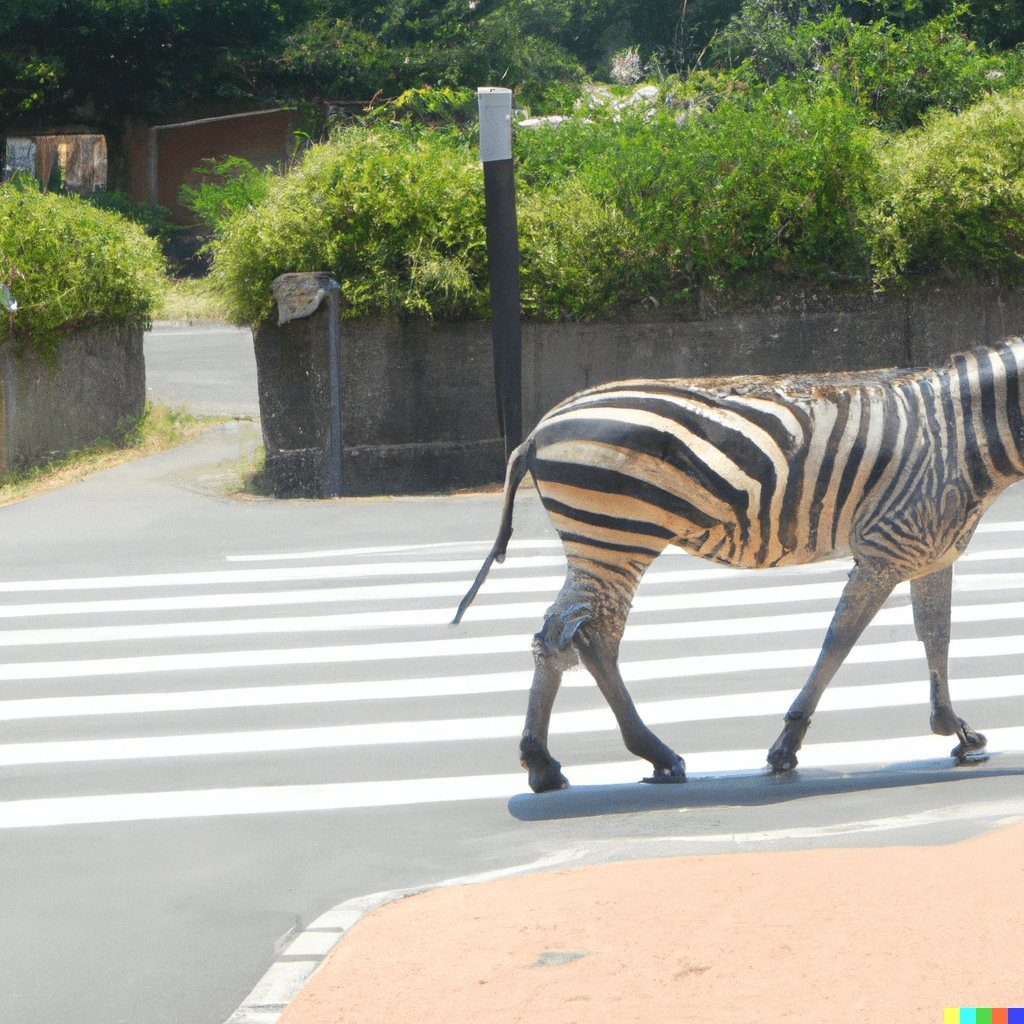}
\includegraphics[width=0.22\textwidth]{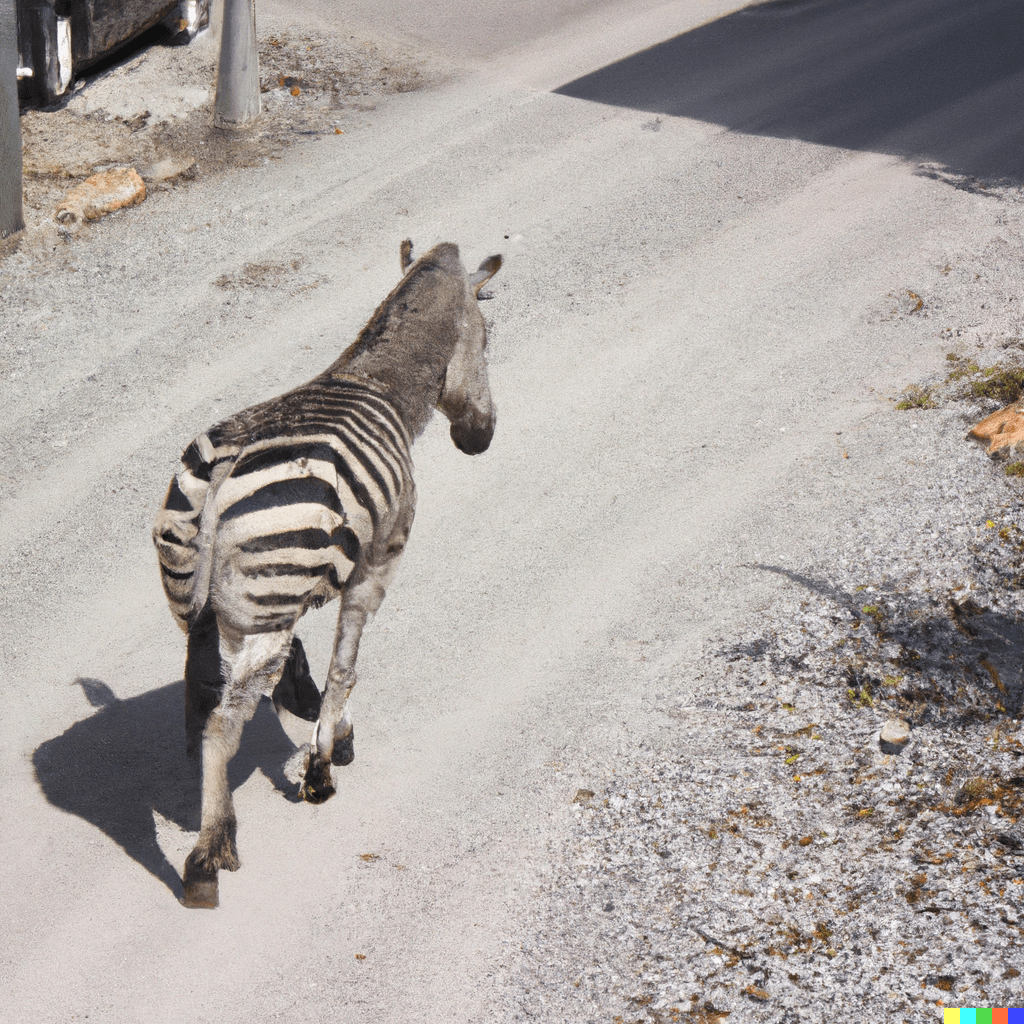}\\
\end{centering}
\noindent\textbf{Main} (left): \emph{A \textbf{\underline{zebra}} and a \textbf{street}}. \\\noindent\textbf{Control} (right): \emph{A zebra and a gravel street}.\\[0.5em]

For the entity-to-property case, we have 10 stimuli-control pairs. On average, the stimuli prompts exhibit the shared property in 92.5\% of the cases while the control prompt shows it in only 6.6\% of the cases.


\indent In similar fashion to the two-properties case, $n_2$ is depicted by \dalle\ with properties from $n_1$, however, this time, $n_1$ is not a property, but an entity. To demonstrate, consider \textit{a zebra and a street}, here, \textit{zebra} is an entity, but it modifies \textit{street}, and \dalle\ constantly generates crosswalks, possibly because of the zebra-stripes' likeness to a crosswalk. And in line with our conjecture, the control \textit{a zebra and a gravel street} specifies a type of street that typically does not have crosswalks, and indeed, all of our control samples for this prompt do not contain a crosswalk. When $n_1$ is a homonym, the entity-to-property case is more nuanced, for example, in \textit{food and a cone hat, zoomed out photo}, the word \textit{cone} modifies \textit{food} by generating ice-cream cones, while its control: \textit{food and a birthday hat, zoomed out photo}, a semantically and physically equivalent replacement of \textit{cone}, does not lead to the generation of ice-cream cone. 

\subsection{Second-order stimuli}
\begin{centering}

\includegraphics[width=0.22\textwidth]{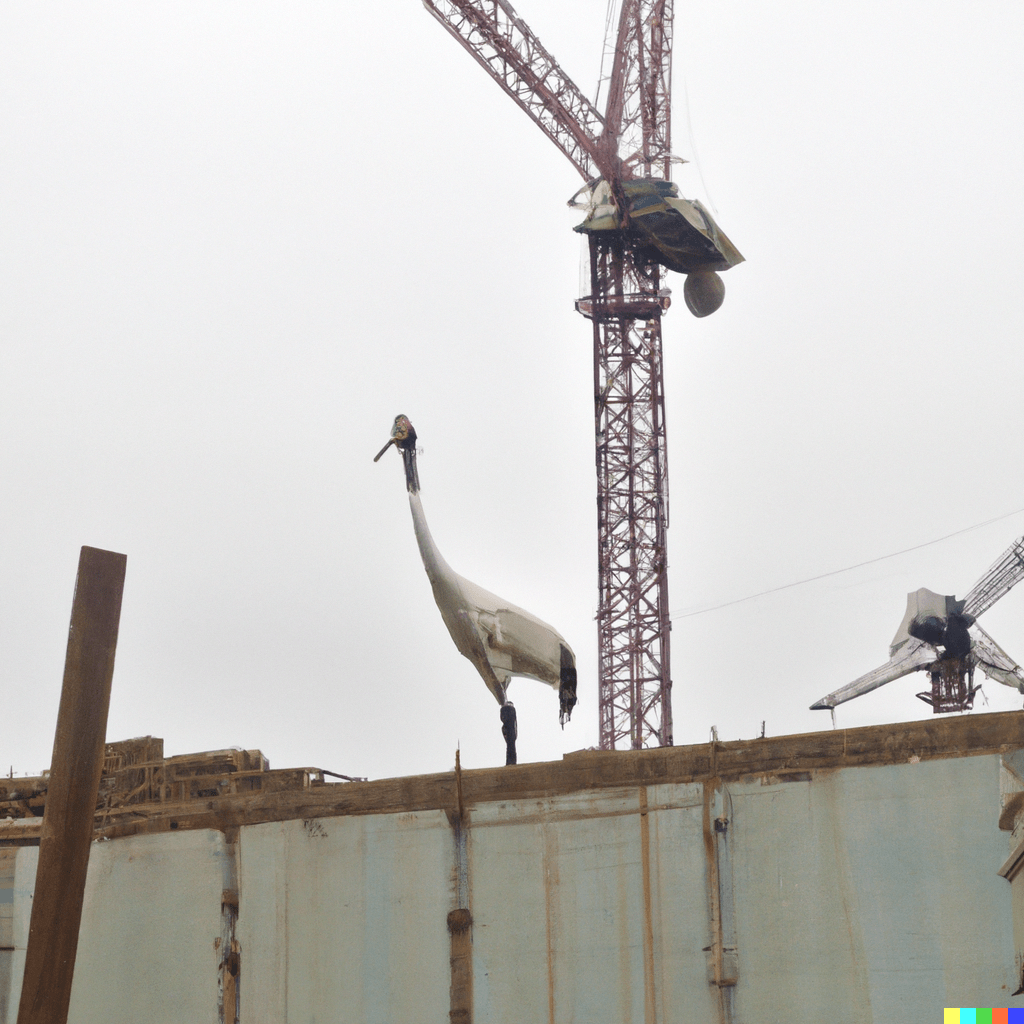}
\includegraphics[width=0.22\textwidth]{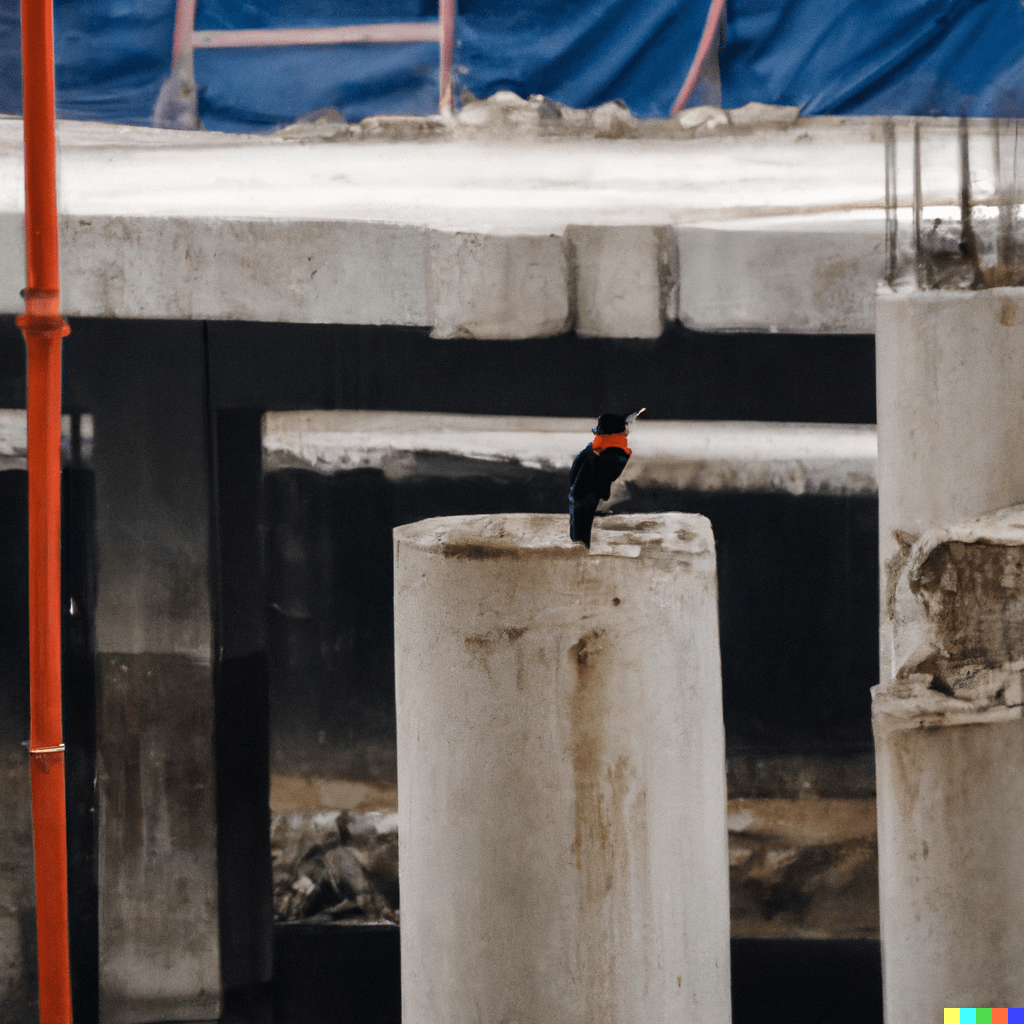}\\
\end{centering}
\noindent\textbf{Main} (left): \emph{a tall, long-legged, long-necked bird and a construction site}. \\\noindent\textbf{Control} (right): \emph{a bird and a construction site}.\\[0.5em]

Concept Leakage can be taken a step further,
by masking the affecting noun and receiving similar results: \textit{a tall, long-legged, long-necked bird and a construction site} is describing \textit{a crane and a construction site}, but nowhere in the prompt a \textit{crane} is mentioned. Yet, it will cause \dalle\ to generate two types of cranes: a bird crane, and a construction crane. We also observe that the bird-crane's head and legs share physical properties with the crane. From the same challenge prompt, the border between a bird-crane and a construction-crane is especially blurred. This behavior is not repeated when the description of a bird-crane is removed from the prompt: \textit{a bird and a construction site}. This suggests that the model first maps entities into a concept space, and only then renders.

\subsection{Other text-to-image-models}
The empirical focus of this work is \dalle. In preliminary experiments, we found that \dallemini\ \citep{dalle_mini}---a much smaller model---did not replicate the observed phenomena. Particularly, only the ``common" word sense is depicted for homonyms. As for \stable\ \citep{stable_diffusion}, the phenomena seem to occur less frequently, particularly because in many times the model does not follow the prompt at all. Other models, such as \citep{imagen, parti}, are not publicly available. We hypothesize that---paradoxically---it is the lower capacity of \dallemini\ and \stable\, and the fact they do not robustly follow the prompts, that make them appear ``better" with respect to the flaws we examine. A thorough evaluation of the relation between scale, model architecture, and concept leakage is left to future work.

\section{Conclusions}

    We demonstrate that across a set of stimuli, \dalle\ does not follow basic principles of symbol-to-entity mapping in language. This flaw is especially pressing given that \dalle\ is directed to the general population, where the majority of the users are probably not aware of the phenomenon.\footnote{While the issues we focus on can easily go unnoticed when examining a randomly-selected set of outputs--which mostly look very good---our examination of carefully-selected stimuli clearly reveals them. We believe that such analysis is important, as human language is characterized by many long-tail phenomena.} Future work should trace back the issues we diagnose to specific components in the model, such as the text encoding or the generative decoder; and study its dependence on scale and architecture. Additionally, the development of post-hoc mitigation techniques is especially important, given that most users lack the resources needed to train the models from scratch.

    
 

    
    
\section{Acknowledgements} 
We thank Carlo Meloni for his valuable feedback.
This project received funding from the Europoean Research Council (ERC) under the Europoean Union's Horizon 2020 research and innovation programme, grant agreement No. 802774 (iEXTRACT).
\section{Limitations}

One limitation of this work is the focus on \dalle\, a commercial product, whose source code is not publicly available. As discussed, capacity issues prevented a thorough examination of the phenomena we focus on in other available models. We hope that additional  models of similar scale will be released to the public in the future, enabling a more thorough examination. Additionally, our analysis is focused on a manually constructed set of stimuli, that were designed so as to surface the issues we focus on in the clearest way. Scaling the analysis would rely on automatic or semi-automatic generation of stimuli of the kind we use, which is a challenging task. Yet, we believe that such large-scale analysis is important in order to robustly quantify the issues we observe.

\bibliography{anthology,references}
\bibliographystyle{acl_natbib}



\clearpage
\newpage

\appendix
\onecolumn
\section{Appendix}

\captionsetup[figure]{labelformat=empty}%
\captionsetup[subfigure]{labelformat=empty}

\subsection{Data}
\label{app:data}
The full image-dataset can be accessed at \url{https://github.com/RoyiRa/DALLE2-Flaws-in-word-to-concept-mapping}. \\

The following tables contain the stimuli we use to test the reference system of \dalle\ . 
\begin{table*}[h]
\centering
\scalebox{0.85}{

\begin{tabular}{l|c}
    \hline
    Prompt & Duplication Ratio \\
    \hline\hline
     A person wearing a \textbf{cone} hat is eating, a full body photo & 7/12  \\
     \hline
     A \textbf{bat} is flying over a baseball stadium & 11/12  \\
     \hline
     A \textbf{bass} lounging in a tropical resort in space, vaporwave & 10/12  \\
     \hline
     A \textbf{fan} at a hot sports event & 12/12  \\
     \hline
     A \textbf{crane} carrying a heavy material, fish in the background & 12/12  \\
     \hline
     A banker collecting \textbf{dough} & 9/12  \\
     \hline
     Two men having a loud \textbf{beef}, restaurant in the background & 10/12  \\
     \hline
     A model with an \textbf{hourglass} figure & 12/12  \\
     \hline
     A gentleman with a \textbf{bow} in the forest & 11/12  \\
     \hline
     A woman is pouring water into her \textbf{glasses} & 9/12  \\
     \hline
     A man stuck in a \textbf{jam}, eating & 7/12  \\
     \hline
     A great \textbf{ruler} & 8/12  \\
     \hline
     \textbf{Apple} commercial & 12/12  \\
     \hline
     A person with \textbf{metal nails} & 6/12  \\
     \hline
     \textbf{date} & 7/12  \\
     \hline
     a person with \textbf{chicken legs}, full body image & 12/12  \\
     \hline
     a \textbf{board} meeting & 9/12  \\
     \hline
    \end{tabular}
    }
    \caption{List of stimuli for the \textbf{multiple entities} case (Section 4.1). The noun that is mapped to two entities is denoted with \textbf{bold}}
    \label{app:homonyms}
\end{table*}


\begin{table*}[h]
\centering
\scalebox{0.85}{
\begin{tabular}{l|c|c}

    \hline
    Challenge and Control & Challenge Ratio & Control Ratio \\
    \hline\hline
    \begin{tabular}{l@{}l@{}} a person \textbf{jay}walking and a \textit{bird} \\ A person crossing the street unlawfully and a bird\end{tabular} & 11/12 & 0/12 \\
     \hline
     \begin{tabular}{l@{}l@{}}\textit{food} and a \textbf{cone} hat, zoomed out photo \\ food and a birthday hat, zoomed out photo\end{tabular} & 12/12 & 0/12 \\
     \hline
      \begin{tabular}{l@{}l@{}}A \textit{fish} and a \textbf{gold} ingot \\ A fish and an ingot\end{tabular} & 12/12 & 0/12 \\
     \hline
     \begin{tabular}{l@{}l@{}}A \textbf{basket}ball near \textit{groceries} \\ A football near groceries\end{tabular} & 12/12 & 5/12\\
    \hline
    \begin{tabular}{l@{}l@{}}A \textbf{lady}bug and a \textit{person}, full body image\\ A bug and a person, full body image\end{tabular} & 11/12 & 6/12\\
     \hline
    \begin{tabular}{l@{}l@{}}An \textit{animal} alongside a \textbf{piggy bank} \\ An animal alongside a table\end{tabular} & 12/12 & 0/12\\
     \hline
    \end{tabular}
    }
    \caption{Challenge and Control Prompts in \textbf{word-to-multiple-modifiers} case (section 4.2). The modifier nouns in the Challenge prompts are denoted with \textbf{bold}, and modified nouns with \textit{italic}}
    \label{app:two_properties}
\end{table*}


\begin{table*}[h]
\centering
\scalebox{0.85}{
\begin{tabular}{l|c|c}

    \hline
    Challenge and Control & Challenge Ratio & Control Ratio \\
    \hline\hline
    \begin{tabular}{l@{}l@{}}A \textbf{zebra} and a \textit{street} \\ A zebra and a gravel street\end{tabular} & 10/12 & 0/12 \\
     \hline
     \begin{tabular}{l@{}l@{}}A \textbf{seal} is opening a \textit{letter} \\ A large water mammal is opening a letter\end{tabular} & 10/12 & 1/12 \\
     \hline
      \begin{tabular}{l@{}l@{}}Two people play \textbf{squash} with \textit{food} \\ Two people play squash\end{tabular} & 12/12 & 0/12 \\
     \hline
     \begin{tabular}{l@{}l@{}}a \textbf{pool} and a \textit{table} \\ a pool and a bar\end{tabular} & 10/12 & 0/12\\
     \hline
    
    \begin{tabular}{l@{}l@{}}A person is washing hair with \textbf{Dove} \textit{shampoo} \\ A person is washing hair with shampoo\end{tabular} & 12/12 & 0/12\\
     \hline
     
    \begin{tabular}{l@{}l@{}}an \textbf{armadillo} and an \emph{elephant} \\ a rhino and an elephant\end{tabular} & 12/12 & 0/12\\
     \hline
    \begin{tabular}{l@{}l@{}}a fish and an \textbf{elephant} \\ a rhino and an elephant \end{tabular} & 12/12 & 0/12\\
     \hline
    \begin{tabular}{l@{}l@{}}A \textbf{cross} and a \textit{sidewalk} \\ A crucifix and a sidewalk\end{tabular} & 10/12 & 2/12\\
     \hline
    \begin{tabular}{l@{}l@{}}classic butter and \textbf{peanuts} \\ classic butter and cucumbers\end{tabular} & 12/12 & 0/12\\
     \hline
    \begin{tabular}{l@{}l@{}}A \textbf{leopard} and a \textit{piece of cloth} \\ a leopard and a black towel \end{tabular} & 11/12 & 5/12\\
     \hline
    \end{tabular}
    }
    \caption{Challenge and Control Prompts in the \textbf{word-to-entity-and-modifier} case (Section 4.3). The modifier nouns in the Challenge prompts are denoted with \textbf{bold}, and modified nouns with \textit{italic}}
    \label{app:object-to-property}
\end{table*}


\begin{table*}[h]
\centering
\scalebox{0.85}{
\begin{tabular}{l|c|c}

    \hline
    Challenge and Control & Challenge Ratio & Control Ratio \\
    \hline\hline
    \begin{tabular}{l@{}l@{}}An \textbf{armadillo} on a \textit{sea} shore \\ A dog on a sea shore\end{tabular} & 12/12 & 0/12 \\
     \hline
     \begin{tabular}{l@{}l@{}}A \textbf{pinniped} is opening a \textit{letter} \\ A large water mammal is opening a letter\end{tabular} & 7/12 & 1/12 \\
     \hline
      \begin{tabular}{l@{}l@{}}a \textbf{tall, long-legged, long-necked bird} and a \textit{construction site} \\ a bird and a construction site\end{tabular} & 12/12 & 0/12 \\
     \hline
     \begin{tabular}{l@{}l@{}}A \textit{person} and a \textbf{bug with red spots}, full body image \\ A bug and a person, full body image\end{tabular} & 11/12 & 6/12\\
     \hline
    \end{tabular}
    }
    \caption{Challenge and Control Prompts in \textbf{Second-Order Concept Leakage} (Section 4.4). The modifier nouns in the Challenge prompts are denoted with \textbf{bold}, and modified nouns with \textit{italic}}
\end{table*}

\clearpage
\newpage

\includepdf[pages=-]{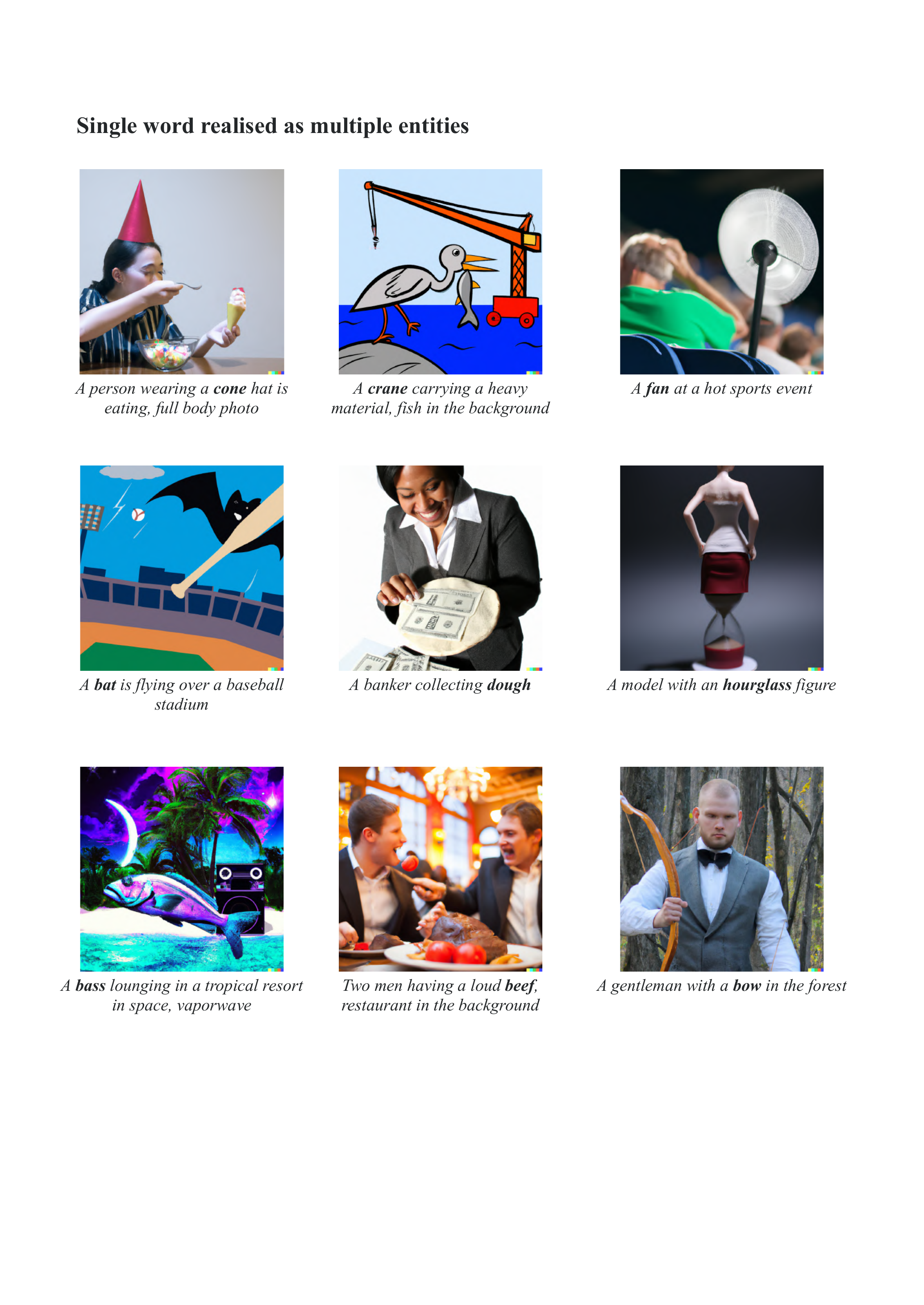}

\end{document}